\pdfoutput=1

\documentclass[10pt,twocolumn,letterpaper]{article}

\usepackage[pagenumbers]{style} 
\usepackage{svg}

\definecolor{BestNIQE}{rgb}{0.7,0.7,0}
\newcommand{\psnr}[1]{\textcolor{red}{#1}}
\newcommand{\lpips}[1]{\textcolor{blue}{#1}}
\newcommand{\niqe}[1]{\textcolor{BestNIQE}{#1}}
\newcommand{\fid}[1]{\textcolor{ForestGreen}{#1}}
\newcommand{\Name}{Linear Combination Diffusion Denoiser} 
\newcommand{\Abkuerzung}{LCDD}

\usepackage{tikz}
\usepackage{float}
\usepackage{algorithm}
\usepackage{algpseudocode}
\usepackage{xcolor}
\usepackage{multirow}
\usepackage{relsize}
\DeclareMathOperator*{\argmin}{arg\,min}
\usepackage{tikz}
\usepackage{todonotes}
\usepackage{float}
\usepackage{xcolor}
\usepackage{colortbl}
\usepackage{booktabs}
\usepackage{setspace}
\usepackage{enumitem}

\usetikzlibrary{automata,positioning}
\usetikzlibrary{patterns}
\usetikzlibrary{calc}
\usetikzlibrary{backgrounds}
\usetikzlibrary{intersections,through,hobby}

\definecolor{specialblue}{HTML}{1f77b4}
\definecolor{specialred}{HTML}{d62728}

\colorlet{discrete}{violet!12}
\colorlet{continuous}{brown!50}
\colorlet{Noise}{green!60}
\colorlet{Data}{black!60}
\colorlet{lincomb}{specialred}

\usepackage[pagebackref,breaklinks,colorlinks,allcolors=specialblue]{hyperref}

\title{ A Simple Combination of Diffusion Models for Better Quality Trade-Offs in Image Denoising }

\author{Jonas Dornbusch\thanks{Currently at Technical University of Munich}\;, Emanuel Pfarr, Florin-Alexandru Vasluianu, Frank Werner, Radu Timofte\\
 University of W\"urzburg\\
{\tt\small  jonas.dornbusch@tum.de}\\  
{\tt\small\{emanuel.pfarr, florin-alexandru.vasluianu, frank.werner, radu.timofte\}@uni-wuerzburg.de}
}

\begin{document}
\maketitle
\begin{abstract}
Diffusion models have garnered considerable interest in computer vision, owing both to their capacity to synthesize photorealistic images and to their proven effectiveness in image reconstruction tasks.
However, existing approaches fail to efficiently balance the high visual quality of diffusion models with the low distortion achieved by previous image reconstruction methods. Specifically, for the fundamental task of additive Gaussian noise removal, we first illustrate an intuitive method for leveraging pretrained diffusion models. Further, we introduce our proposed \Name$\;$(\Abkuerzung), which unifies two complementary inference procedures—one that leverages the model’s generative potential and another that ensures faithful signal recovery. By exploiting the inherent structure of the denoising samples, \Abkuerzung$\;$achieves state-of-the-art performance and offers controlled, well-behaved trade-offs through a simple scalar hyperparameter adjustment.

\end{abstract}    
\section{Introduction}
\label{sec:intro}
The advances in camera technology over the past decades have enabled a large number of people worldwide to carry a camera at all times, leading to an uncountable number of images being taken every day.
Unfortunately, several factors can limit the ability of digital imaging to accurately capture reality, such as rapid camera movement and imperfect or damaged sensors.  
These flaws are addressed by the field of image reconstruction, which aims to recover the true image from a corrupted source and encompasses tasks such as super-resolution \cite{Sahak_SR3+, saharia2021image}, inpainting \cite{RePaint}, and deblurring \cite{whang2022_deblurring_stochastic_refinement}.  
A particularly important example of image reconstruction is denoising, which aims to remove random noise present in an image \cite{Elad2023ImageDenoisingSurvey}.

\begin{figure}[tp]
\centering
    \resizebox{0.5\textwidth}{!}{\begin{tikzpicture}[
        dot/.style = {circle, fill, minimum size=#1,
            inner sep=0pt, outer sep=0pt},
        ]
            \node[]                 (input)                      []                 {\includegraphics[width=.05\textwidth]{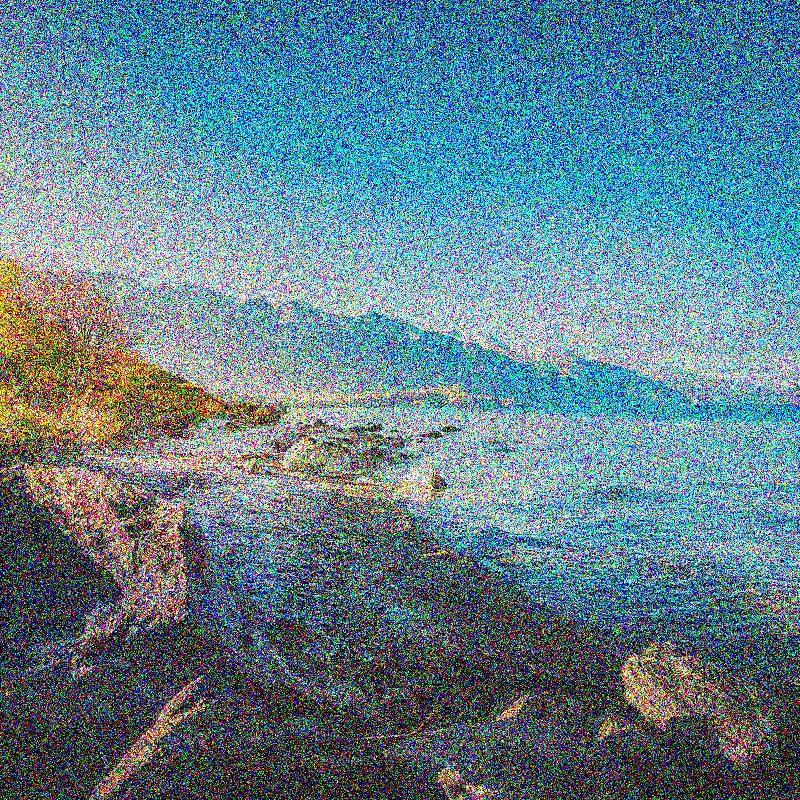}};
            \node[]                 ()                      [below=0cm of input]   {\textbf{Input}};
            \node[]                 (scaled)                      [right=1cm of input]                 {\includegraphics[width=.05\textwidth]{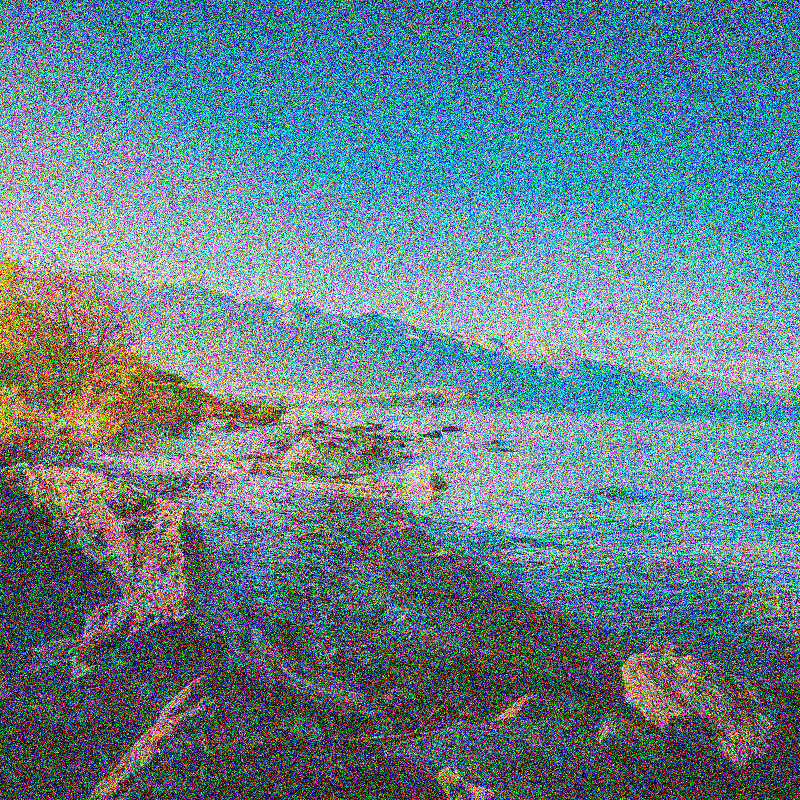}};
            \draw[->, shorten <=-0.4mm, shorten >=-0.4mm]  (input.east) -- (scaled.west)
        node[midway, above=0cm, xshift=0.8mm] ()  {\footnotesize{$\cdot \hat \alpha$}} node[midway, below=0cm, xshift=0.8mm] ()  {\footnotesize{scaling}};

        \node[]                 (x5c)            [above right=0.9cm and 0.1cm of input]      {\includegraphics[width=.05\textwidth]{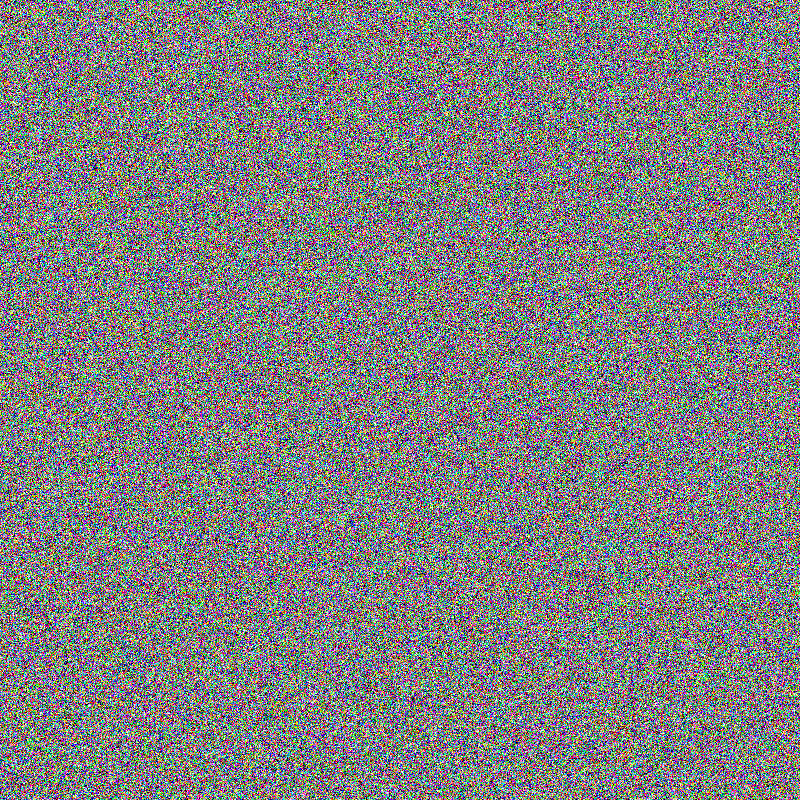}};
        \node[dot=0.1cm] (circ1c) [right=0.05cm of x5c]  {};
        \node[dot=0.1cm] (circ2c) [right=0.05cm of circ1c]  {};
        \node[dot=0.1cm] (circ3c) [right=0.05cm of circ2c]  {};
        \node[]                 (x3c)                      [right=0.05cm of circ3c]                 {\includegraphics[width=.05\textwidth]{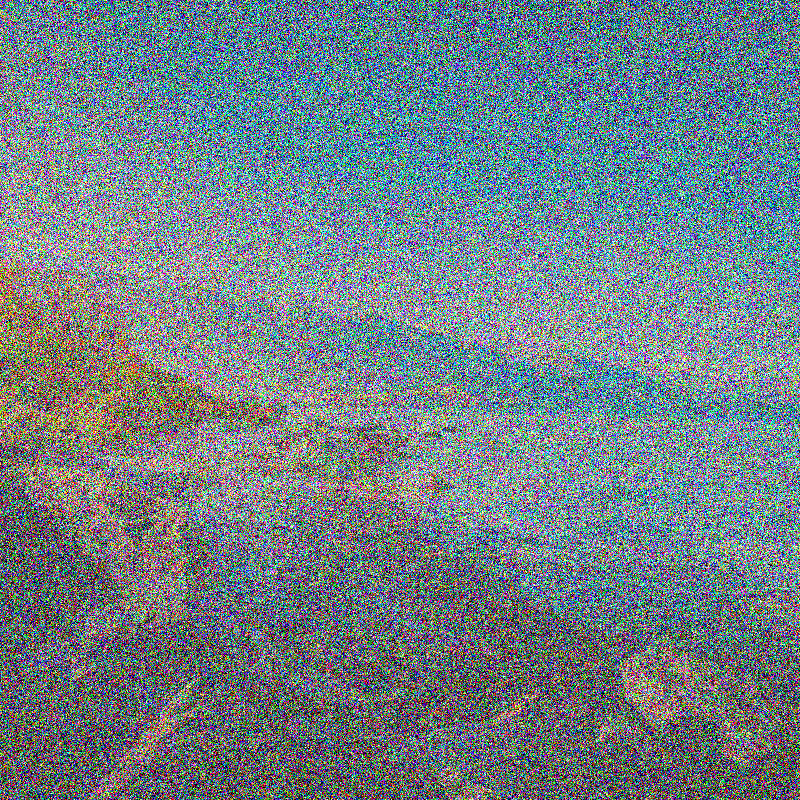}};
      \node[]                 (x2c)                      [right=0cm of x3c]  {\includegraphics[width=.05\textwidth]{ExampleImage2Noise}};
      \node[]                 (x1c)                      [right=0cm of x2c]  {\includegraphics[width=.05\textwidth]{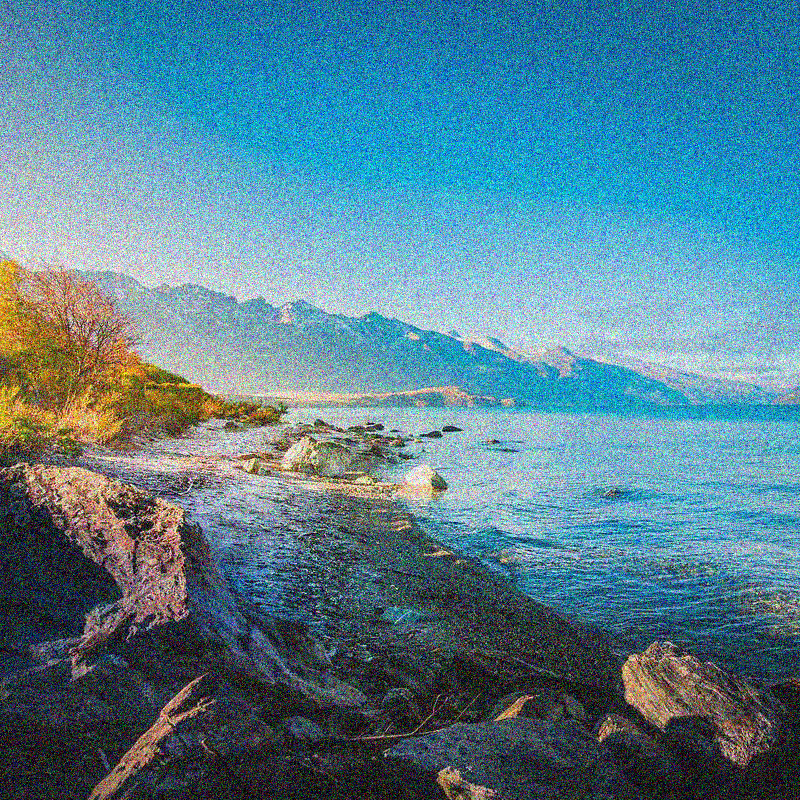}}; 
        \node[dot=0.1cm] (circ1c) [right=0.05cm of x1c]  {};
        \node[dot=0.1cm] (circ2c) [right=0.05cm of circ1c]  {};
        \node[dot=0.1cm] (circ3c) [right=0.05cm of circ2c]  {};
      \node[]                 (x0c)                      [right=0.05cm of circ3c ]   {\includegraphics[width=.05\textwidth]{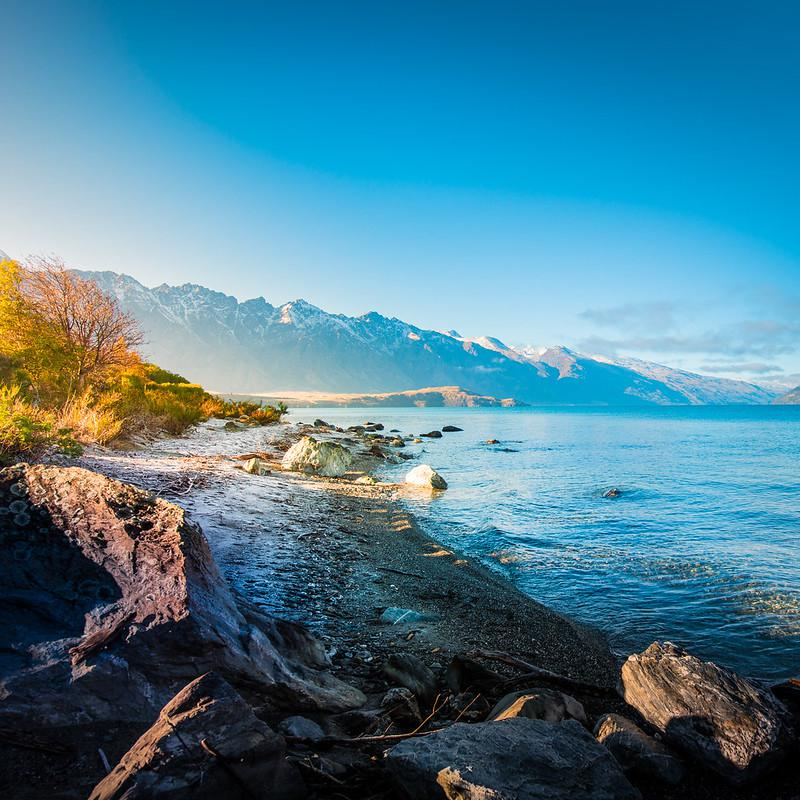}};

        \draw[->, thick, dashed, teal, shorten >=2pt, shorten <=3pt]
        let 
            \p1 = (x2c),
            \p2 = (x1c),
            \p3 = (circ2c),
            \p4 = (x0c)
        in 
            ($(\x1, \y1)+(0,-0.5)$) to [out = -60, in = -120, looseness = 0.7] ($(\x2, \y1)+(0,-0.5)$); 
         \draw[->, thick, dashed, teal, shorten >=2pt, shorten <=3pt]
        let 
            \p1 = (x2c),
            \p2 = (x1c),
            \p3 = (circ2c),
            \p4 = (x0c)
        in 
             ($(\x2, \y1)+(0,-0.5)$) to [out = -60, in = -120, looseness = 1.0] ($(\x3, \y1)+(0,-0.5)$);
         \draw[->, thick, dashed, teal, shorten >=2pt, shorten <=3pt]
        let 
            \p1 = (x2c),
            \p2 = (x1c),
            \p3 = (circ2c),
            \p4 = (x0c)
        in   
            ($(\x3, \y1)+(0,-0.5)$) to [out = -60, in = -120, looseness = 1.0] ($(\x4, \y1)+(0,-0.5)$);

      \draw[->, thick] (scaled.east) to[out=0, in=-120] node[pos=0.65, fill=white, rectangle, draw=black, thin, xshift=-6pt] {\footnotesize{Insert in step $\hat k$}} (x2c.south);

        \node[]                 (x5o)                      [below right=0.5cm and -0.4cm of input]                 {};
        
      \draw[]
        let 
            \p1 = (x0c),
            \p2 = (x5o)
        in 
            ( $(\x1, \y2)$) node[] (x0o) {\includegraphics[width=.05\textwidth]{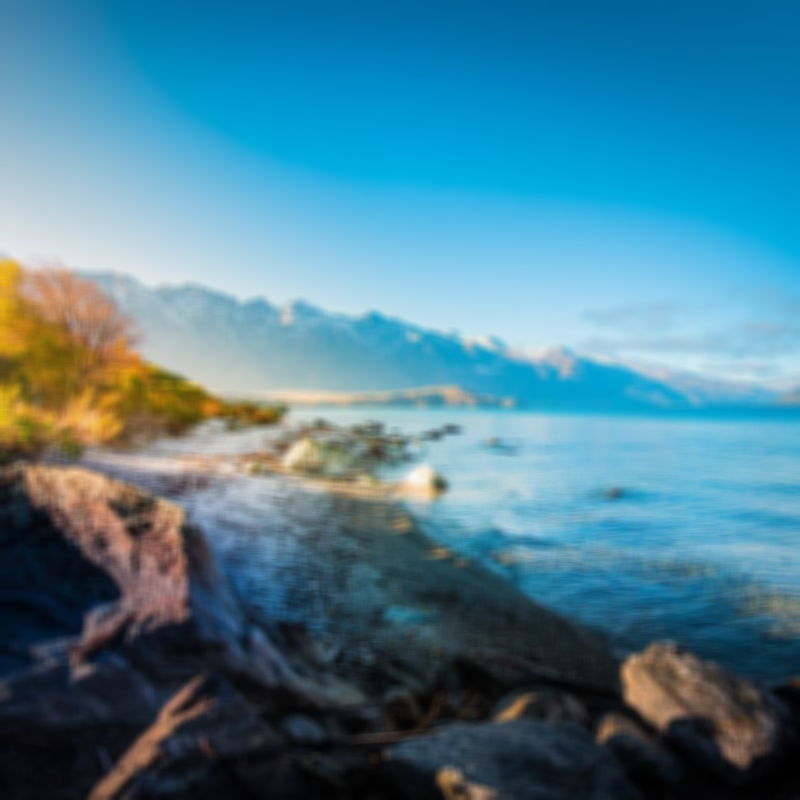}};

        \shade[top color=blue!30, bottom color=green!30, rounded corners=5pt, draw=black, thick]
            let 
                \p1 = (x0c.east),
                \p2 = (x0o.west)
            in
                ($(\x1, \y1)+(0.3,0.3)$) rectangle ($(\x2, \y2)+(3,-0.3)$);

        \node[align=center, font=\scriptsize\bfseries] 
            at ($($(x0c.east)$)+(1.05,-0.05)$) {
                High Visual \\ Quality
            };
        \node[align=center, font=\scriptsize\bfseries] 
            at ($($(x0o.east)$)+(1.05,0)$) {
                Low \\ Distortion
            };

        \draw[<->, thick]  ($($(x0c.east)$)+(2.05,-0.05)$) to node[midway,rotate=90, anchor=south, yshift=-14pt] {\footnotesize{controllable by $\lambda$}} ($($(x0o.east)$)+(2.05,0)$);
        
    \path (x2c.south) to[out = -60, in = -120, looseness = 0.7] node[pos=0.5] (midpoint_update) {} (x1c.south);
    \draw[->, thick] (midpoint_update) to[out=-90, in=180] node[midway, fill=white, rectangle, draw=black, thin] {\footnotesize{Initial Prediction}} (x0o.west);

      \draw[-, lincomb, thick]  (x0c.south) -- (x0o.north) node[midway,rotate=90, anchor=south, text width=2cm, align=center] (midLC)  {\footnotesize{\raisebox{-5pt}{linear} \\ combination}};
      \draw[]
        let 
            \p1 = (input),
            \p2 = (midLC)
        in 
            ( $(\x2, \y1)+(2.02,0.4)$) node[] (output)  {\includegraphics[width=.05\textwidth]{ExampleImage0Noise}};
        \draw[->,lincomb, thick]
        let 
            \p1 = (output),
            \p2 = (midLC)
        in 
            ( $(\x2, \y1)+(0.35,0)$) -- (output.west);
      \node[]                 ()                      [below=0cm of output]  
            {\textbf{Output}};

        \begin{scope}[on background layer]
        
        \draw[rounded corners=5pt, dashed]
        let 
            \p1 = (x5c),
            \p2 = (x0o.east)
        in
            ($ (\x1, \y1)+ (-0.6,1.5) $) rectangle($ (\x2, \y2) + (0.1,-0.55) $); 
        \draw[fill=continuous, rounded corners=5pt]
        let 
            \p1 = (x5c),
            \p2 = (x0c)
        in
            ($ (\x1, \y1)+ (-0.5,-0.82) $) rectangle($ (\x2, \y2) + (0.55,1.4) $) 
            ($ (\x1, \y1)+ (-0.5,1.4) $) node[anchor = north west] {\footnotesize{Classical Diffusion}};
        \draw[fill=black!10, rounded corners=5pt]
        let 
            \p1 = (x5c),
            \p2 = (x3c)
        in
            ($ (\x1, \y1)+ (-0.5,-0.55) $) rectangle($ (\x2, \y2) + (0.5,0.9) $) 
            ($ (\x1, \y1)+ (-0.5,0.9) $) node[anchor = north west] {\footnotesize{unused}};

        \end{scope}

        \node[anchor=north east, xshift=8pt, yshift=5pt] at (current bounding box.north east) {
            \begin{tikzpicture}[baseline=-0.5ex]
                \draw[->, thick, dashed, teal, shorten >=1pt, shorten <=1pt] (0, 0) -- (0.5cm, 0);
                \node[anchor=west] at (0.3cm, 0) {\tiny\textbf{\begin{tabular}{c} Diffusion \\ Update \end{tabular}}};
            \end{tikzpicture}
        };
    \end{tikzpicture}}
    \caption{Overwiev of our proposed \Abkuerzung$\;$ method. A noisy image is scaled, such that it closely resembles an intermediate step in the inference process of a classical DM, and denoised using the structure of DMs. The denoised image is then linearly combined with the initial prediction of the first diffusion step, allowing us to achieve favorable distortion-perception trade-offs.}
    \label{fig:MethodFigure}
\end{figure}

The performance of image reconstruction methods is commonly measured in two different metrics: 
One metric measures the similarity of the reconstruction to the ground truth, often referred to as \textit{distortion}, while another quantifies how natural the image appears to the human eye, often termed \textit{perception}.  
Blau and Michaeli \cite{Blau2018_perception_distortion_tradeoff} proved theoretical limits on optimizing both performance measures simultaneously, introducing the distortion-perception trade-off. 
Further theoretical work on this trade-off was done by Freirich \etal \cite{Freirich2021_distortion-perception_tradeoff_wasserstein}, who argued that a simple linear combination of the outputs from a reconstruction method optimized for distortion and another optimized for perception should yield a prediction with an optimal trade-off between the two measures. However, practical explorations of the trade-off, such as \cite{adrai2024DeepOptimal,ohayon2024posteriormeanrectifiedflowminimum,whang2022_deblurring_stochastic_refinement,wu2024detail_aware_denoising} reveal discrepancies from the theoretical predictions, \ie they do not show optimal or even advantageous trade-offs.

In the past, various approaches have been proposed to address the denoising problem, ranging from mathematical techniques \cite{Buades_NonLocalMeanDenosing2005} to machine and deep learning-based methods \cite{Elad_DeepLearningDenoisng2006,zhang2017beyond,liang2021_swinir, dabov2006image, Guo2024_MambaIR, Zamir2021_Restormer, zhang2020plug}. 
Traditionally, machine learning-based denoising methods aim to minimize distortion, leading to strong performance in this aspect. 
However, they often struggle with perceptual quality, producing images that appear overly smoothed and unnatural.  
Recently, the success of diffusion models (DMs) \cite{Ho2020_DDPM,song2021_DDIM} has led to their adoption in image denoising. 
Given their ability to generate high-quality images, they are particularly appealing when maintaining a natural appearance is a priority. 
Denoising methods utilizing DMs include the work of Xie \etal \cite{xie2023diffusionmodelgenerativeimage}, who expensively train a specially designed diffusion model, and Wu \etal \cite{wu2024detail_aware_denoising}, who apply the computationally expensive full inference process of a diffusion model in the residual space. 
While both approaches achieve strong perceptual quality, they exhibit weaker performance in terms of distortion.

\begin{figure}[tp]
\centering
    \resizebox{0.25\textwidth}{!}{\begin{tikzpicture}
        \draw (0,2) node[anchor=south, rotate=90] () {\footnotesize Distortion↑};
        \draw (2,0) node[anchor=north] () {\footnotesize Perception↓ };
        \fill [fill=red!5]
         (0,0) node              {}
      -- (4,4) node[behind path] {}
      -- (4,0) node              {};
        \fill [fill=blue!5]
         (0,0) node              {}
      -- (4,4) node[behind path] {}
      -- (0,4) node              {};
      \filldraw [black] (0.5,0.5) circle (1pt);
      \draw (0.5,0.5) node[anchor=north] () {\footnotesize Method 1};
      \filldraw [black] (3.5,3.5) circle (1pt);
      \draw (3.5,3.5) node[anchor=north] () {\footnotesize Method 2};
      \draw[green!80] (3.5,3.5) to[out=180,in=30] (2,3);
      \draw[green!80] (2,3) to[out=210,in=80] (0.5,0.5);

      \draw[red] (3.5,3.5) to[out=180,in=70] (2,2.5);
      \draw[red] (2,2.5) to[out=260,in=20] (1,0.7);
      \draw[red] (1,0.7) to[out=200,in=0] (0.5,0.5);
      
      \draw[red] (3.5,3.5) to[out=260,in=30] (2,1);
      \draw[red] (2,1) to[out=210,in=20] (0.5,0.5);
      \draw[->] (0,0) to (4.0,0); 
        \draw[->] (0,0) to (0,4.0); 
    \end{tikzpicture}}
    \caption{The trajectories of three methods compared to two reference methods. The distortion measure is designed such that higher scores are better, while for the perception measure lower scores are better. The method that describes the green trajectory has consistently advantageous trade-offs, while the methods describing either of the red trajectories do not.}
    \label{fig:TradeOffBeispiel}
\end{figure}
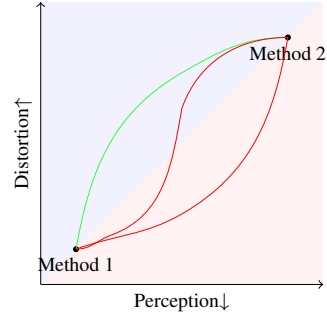

To bridge the gap between distortion and perception, we propose the \Name$\;$(\Abkuerzung), a denoising scheme that leverages the inherent iterative denoising structure of diffusion models. 
Our approach involves inserting a noisy image at an appropriate step in the inference process of pretrained diffusion models and linearly combining two different inference outputs to achieve a controlled and advantageous distortion-perception trade-off.  
For our proposed linear combination, we use the fact that our way of using DMs for denoising naturally exhibits a distortion-perception trade-off when varying the length of the inference schedule, where we achieve strong distortion results for shorter schedules, while longer schedules sacrifice distortion in favor of improved perception. 
By linearly combining a distortion-focused one-step schedule with a perception-focused multi-step schedule, we obtain a distortion-perception trade-off that is consistently advantageous, \ie it is strictly better than a linear trade-off (\cf \cref{fig:TradeOffBeispiel}), regardless of the combination factor. 
Moreover, our approach not only outperforms the natural trade-off but even surpasses the original reconstructions in both distortion and perception in some cases. Additionally, it capitalizes on the expressiveness of pretrained diffusion models, avoiding the need for further training while enabling denoising across all noise levels with a single model.
Interestingly, this also serves as a further practical study into the theory of Friedrich \etal \cite{Freirich2021_distortion-perception_tradeoff_wasserstein}.

\section{Related work}
\label{sec:relatedWork}
\textbf{Denoising:} The field of denoising is well explored and dacedes old \cite{Elad2023ImageDenoisingSurvey}. Initial methods for suppressing or estimating noise in images were of mathematical nature, such as Buades \etal \cite{Buades_NonLocalMeanDenosing2005}, or Dette \etal \cite{Dette1998_variance_estimation} and Gasser \etal \cite{gasser1986_variance_estimation}.
After Chatterjee and Milanfar \cite{DenosingIsDead} famously declared that "Denoising is Dead", the success of machine and deep learning across various fields led to their adoption in denoising, revitalizing progress in the field. 
The first competitive machine learning-based denoising algorithm was the K-SVD algorithm, introduced by Elad and Aharon \cite{Elad_DeepLearningDenoisng2006}. 
Notable advancements followed, including the DnCNN network by Zhang \etal \cite{zhang2017beyond, zhang2020plug}, SwinIR by Liang \etal \cite{liang2021_swinir}, and BM3D by Dabov \etal \cite{dabov2006image}, all of which significantly surpassed traditional approaches. 
Methods currently representing the state of the art in denoising include the Restormer network by Zamir \etal \cite{Zamir2021_Restormer}, and MambaIR by Guo \etal \cite{Guo2024_MambaIR}.

\textbf{Diffusion models:} After their first appearance in the work of Sohl-Dickstein \etal \cite{sohl-dickstein2015_og_diffusion}, DMs gainend considerable traction with the work of Ho \etal \cite{Ho2020_DDPM} (DDPM) and Song \etal \cite{song2021_DDIM} (DDIM). 
The theory and practical capabilities of DMs have since been refined considerably, \cf \cite{Dhariwal2021_diff_beat_gan,Song2020ScoreBasedGM_SDE,Huang2021_score_objective_equivalent,nichol21_improving_ddpm,song2019_score_matching}.
While diffusion models show remarkable capabilities in image generation, \eg shown in Stable Diffusion by Rombach \etal \cite{rombach2022high}, several approaches have been proposed to leverage diffusion models in image restoration tasks \cite{li2023_diffusion_image_restoration_survey}, such as super-resolution \cite{saharia2021image,Sahak_SR3+, zhu2023_diffpir}, deblurring \cite{zhu2023_diffpir}, inpainting \cite{RePaint, zhu2023_diffpir} or general inverse problems \cite{kawar2021snips,kawar2022denoising,song2021solving}.

Previous work on diffusion models for denoising include Xie \etal\cite{xie2023diffusionmodelgenerativeimage}, who rely on a specifically trained model, and Wu \etal \cite{wu2024detail_aware_denoising}, who use full diffusion inference in the residual space to denoise images.
Further examples include \cite{manor2024DenosingWithDiffusion1,manor2024DenosingWithDiffusion2,Wang2023ReconstructandGenerateDM}.

\textbf{Distortion-perception trade-off:} Ohayon \etal \cite{ohayon2024posteriormeanrectifiedflowminimum} propose optimizing for MSE under the constraint of perfect perception. 
Their approach utilizes an initial prediction, followed by the application of a rectified linear flow to guide the prediction toward the data distribution. 
Wang \etal \cite{Wang2023ReconstructandGenerateDM} apply diffusion models to image patches, investigating the impact of different inference schedules on perceptual quality. 
Whang \etal \cite{whang2022_deblurring_stochastic_refinement} present a method that navigates the trade-off between distortion and perception in image deblurring. 
Their approach involves generating a low-distortion initial prediction and performing diffusion in the residual space. 
The work of Wu \etal \cite{wu2024detail_aware_denoising} also explores the trade-off in a similar manner, particularly in the context of denoising.

Theoretical work on the trade-off between distortion and perception was first introduced by Blau and Michaeli \cite{Blau2018_perception_distortion_tradeoff}. 
Further exploration within the context of Wasserstein space was conducted by Freirich \etal \cite{Freirich2021_distortion-perception_tradeoff_wasserstein}, who demonstrated that a simple linear combination of two estimators, each lying on the optimal distortion-perception trade-off curve, results in an estimator that also resides on this optimal curve. 
Building on this theory, Adrai \etal \cite{adrai2024DeepOptimal} proposed a method that takes a low-distortion estimator and estimates the optimal transport plan to map it onto the data distribution. 
According to the theory, the resulting transported estimator is visually optimal and remains on the optimal distortion-perception trade-off curve. 
However, the predicted behavior of the linear combination could not be empirically observed.

\section{Background}
\label{sec:DiffusionModels}

DDPM \cite{Ho2020_DDPM} operates by progressively corrupting a natural image into pure noise over multiple discrete steps and learning the reverse process to reconstruct a natural-looking image from pure noise. Consequently, a diffusion model comprises multiple denoisers, each corresponding to a specific corruption step. The corruption process, starting from a clean image $x_0$, can be expressed by the update step
\begin{equation}
    \label{eq:ForwardProcessAlpha}
     x_k = \sqrt{\bar\alpha
     _k} x_{0} + \sqrt{1-\bar\alpha_k}\eta_k, \quad \eta_k\sim \mathcal{N}(0,I_d),
\end{equation}
with 
\begin{equation}
    \alpha_k \in [0,1], \quad \text{and} \quad \bar \alpha_k=\prod_{i=1}^k \alpha_k,
\end{equation}
which shows that the conditional forward marginals $q_{k|0}$ are Gaussian. The generating procedure works by training a neural network $\eta_\theta(x_k,k)$ to predict the starting point $x_0$ of the Markov chain containing $x_k$, and then sampling from the tractable quantity $q_{k-1|k,0}(x_{k-1}|x_k, \eta_\theta(x_k,k))$, which is the density of the $(k-1)$-step of the forward process, conditioned on updated iterative $x_k$, and the starting point $x_0$. This results in the inference update step 
\begin{equation}
    \label{eq:DDPMupdate}
    x_{k-1} = \mu_\theta(x_k,k) + \sigma_k \varepsilon_k, 
\end{equation}
where 
\begin{align}
    \mu_\theta(x_k,k)&=\frac{1}{\sqrt{\alpha_k}} \left(x_k - \frac{1-\alpha_k}{\sqrt{1-\Bar{\alpha}_k}}\eta_\theta (x_k,k)\right),\\
    \varepsilon_k &\sim \mathcal{N}(0,I_d),
\end{align}
and $\sigma_k$ being chosen in advance.
Building upon this formulation, DDIM \cite{song2021_DDIM} consists of a forward process that has the same forward marginals $q_{k|0}$, but of a different backward process, and thus a different generating scheme. It is given as 
\begin{equation}
    \label{eq:DDIMupdate}
    x_{k-1}= \mu_{x_k}x_k + \mu_{x_0}\eta_\theta(x_k,k),
\end{equation}
with \begin{equation}
\mu_{x_k}=\frac{\sqrt{\bar\alpha_{k-1} }}{\sqrt{\bar\alpha_{k} }}
\end{equation}
and 
\begin{equation}
    \mu_{x_0}=\left(\sqrt{1-\bar \alpha_{k-1}}-\frac{\sqrt{\bar\alpha_{k-1} }}{\sqrt{\bar\alpha_{k} }}\sqrt{1-\bar \alpha_k}\right).
\end{equation}
Since the training procedure of DDPM only uses the representation of the forward marginals $q_{k|0}$, the same network $\eta_\theta$ can be used for inference in DDIM, since the forward marginals of the two match.\\

\section{Approach}
\label{sec:Approach}
For clarity, we divide this section into three parts. The Full Method is depicted in \cref{fig:MethodFigure}  
\subsection{Denoising with diffusion models}
\label{subsec:DenosiingWithDiffusionModels}
As discussed in the previous section, diffusion models generate images through iterative denoising. If we can identify a suitable time step at which our noisy image closely resembles the corresponding latent variable, we can insert it into the diffusion model at that point and leverage the pretrained model for denoising. However, the structure of noisy images may differ depending on the forward process. In standard denoising, an image is given as  
\begin{equation}
    y = x_0 + \rho n,
\end{equation}  
where $n$ is additive Gaussian white noise. In contrast, in DDPM \eqref{eq:ForwardProcessAlpha}, the clean image $x_0$ is scaled by a factor $\sqrt{\bar \alpha_k}\neq 1$. To bridge this gap, we scale the noisy image $y$ by $\sqrt{\hat{\alpha}}$, resulting in  
\begin{equation}
    \sqrt{\hat{\alpha}} y = \sqrt{\hat{\alpha}} x_0 + \sqrt{\hat{\alpha}} \rho n.
\end{equation}  
Now, if the conditions 
\begin{equation}
\label{eq:DenoisingCondition}
    \sqrt{1-\hat \alpha} \stackrel{!}{=} \sqrt{\hat \alpha}\rho,
\end{equation}
and 
\begin{equation}
    \hat{k} = \argmin_{k=1,2,...,N} |\bar{\alpha}_k - \hat{\alpha}|
\end{equation}
are met, the noisy image $\sqrt{\hat\alpha}y$ closely resembles the $\hat k$-th latent variable of DDPM or DDIM. In practice, we determine $\hat k$ by solving \eqref{eq:DenoisingCondition}, which yields \begin{equation}
    \hat{\alpha} = \frac{1}{1+\rho^2},
\end{equation} 
and then search the noise levels $\bar\alpha_k$ after the one that closeset matches $\hat\alpha$. Finally, we insert the image $\sqrt{\hat\alpha}y$ in the resulting step $\hat k$ to obtain clean and denoised outputs.

Although a small discrepancy may exist between $\hat{\alpha}$ and $\bar{\alpha}_k$, DDPM is trained with $N=1000$ steps, making the difference $|\bar{\alpha}_{\hat{k}} - \hat{\alpha}|$ negligible. Since the forward marginals $q_{k|0}$ are identical for DDPM and DDIM, this approach applies equally to both models. We outline the complete denoising procedure in \cref{alg:DenoisingWithDiffusion}. Notably, the algorithm primarily consists of the standard DDPM and DDIM inference updates, as defined in \eqref{eq:DDPMupdate} and \eqref{eq:DDIMupdate}.

\begin{algorithm}[h]
    \centering
    \caption{Denoising with diffusion models}
    \label{alg:DenoisingWithDiffusion}
    \textbf{Input:} Noisy image $y$, noise level $\rho$, trained model $\eta_\theta$, noise schedule $\bar\alpha_1,...,\bar\alpha_N$, sampling variances $\sigma_2,...,\sigma_N$
    \begin{algorithmic}[1]
        \State $\bar \alpha_0 \gets 1$
        \State $\hat{\alpha} \gets \frac{1}{1+\rho ^2}$
        \State $\hat k \gets \argmin_{k=1,2,...,N}|\bar\alpha_k - \hat{\alpha}|$
        \State $x_{\hat k}\gets \sqrt{\hat{\alpha}} \, y$
        \State$\textbf{for}\; t = \hat k,...,1 \;\textbf{do} $
        \State $\quad \textbf{if} \; \text{DDPM}$
        \State  $\quad \quad  z \sim \mathcal{N}(0,I_d)$  
        \State $\quad \quad \textbf{if} \; t=1 \; \textbf{then} \; z\gets 0$
        \State$\quad \quad x_{t-1} \gets \mu_\theta (x_t,t) + \sigma_tz$  
        \State $\quad \textbf{if}\; \text{DDIM}$
        \State  $ \quad \quad x_{t-1}\gets \mu_{x_t}x_t + \mu_{x_0}\eta_\theta(x_t,t)$
        \State $\quad \textbf{end if}$
        \State $\textbf{end for}$
        \State $\textbf{return} \; x_0$ 
    \end{algorithmic}
\end{algorithm}

 \cref{alg:DenoisingWithDiffusion} reviles an additional advantage of our proposed denoising method: we only use one model, $\eta_\theta$, for denoising on all noise levels $\rho$, while other denoisers based on neural networks , \eg \cite{Zhang2017DnCNN,Zamir2021_Restormer,Guo2024_MambaIR} have to train and employ different networks for different noise scales to get optimal results. Furthermore, it clarifies that our proposed denoising method requires at most $\hat{k} \leq N$ steps. Depending on the noise level, this results in a considerably shorter inference process, reducing the number of neural function evaluations (NFEs) compared to standard DDPM and DDIM inference.
 
 Using noise estimators \cite{Dette1998_variance_estimation,gasser1986_variance_estimation}, the variance $\rho^2$ of a noisy image $y = x_0 + \rho n$, where the noise follows a Gaussian distribution $n \sim \mathcal{N}(0, I_d)$, can be estimated with high accuracy. By first applying such an estimator to obtain an estimate of $\rho$ for a noisy image $y$ with Gaussian noise of unknown variance, our method can be adapted into a blind denoising approach.

\subsection{Varying inference schedules}
\label{subsec:VaryingScheduleLength}
We vary the length of the inference process schedule in a similar way to DDIM: given the noise level $\rho$ of the noisy image $y = x_0+\rho n$ with $n \sim \mathcal{N}(0,I)$, the optimal time step $\hat k$ for denoising is determined as described in the previous section. While the original diffusion model performs denoising at every step $k=1,...,\hat k$, we select a subset $\tau$ of $\{1,2,..., \hat k\}$. Specifically, let $\tau$ be a sequence of elements in $\{1,2,...,\hat k\}$ with $\tau_i < \tau_{i+1}$ for $i =1,2,...,|\tau|-1$, and $\hat k\in  \tau$. A new noise schedule $\alpha^\text{new}$ is then created by defining
 \begin{equation}
    \alpha_i^\text{new}=\begin{cases}
    \alpha_{\hat k}, &\text{if} \; i= |\tau|\\
    \prod_{t=\tau_i}^{\tau_{i+1}} \alpha_t, & \text{otherwise}
        \end{cases}
 \end{equation}
 for $i=1,2,...,|\tau|$. With this choice, we ensure that $\alpha^\text{new}_i=\bar \alpha_{\tau_i}$, preserving the forward
 process while using fewer latent variables. If we choose to perform $\hat N$ steps in the reverse denoising starting from step $\hat k$, we can distribute these steps evenly across $\{1,2,...,\hat k\}$, to get a new equidistant inference schedule $\tau$ of length $\hat N$. 

 \subsection{Linearly combining images}
 A simple yet powerful approach to generate trade-offs is to take one image $I_D$ with low distortion and another image $I_{P}$ with high perceptual quality, then perform a linear combination 
 \begin{equation}
     LC_\lambda=\lambda I_D + (1-\lambda)I_{P}, \quad \lambda \in  [0,1]
 \end{equation}
 of the two. For our method we take the low distortion image to be the one generated through a one-step schedule, while the high perceptual quality image is generated through a multi-step schedule. While it may seem that this method would lead to suboptimal perception and distortion scores, the results are remarkably strong, even surpassing the initial denoised images in both perception and distortion, as we will show in \cref{sec:ExperimentalResults}. Consequently, our full method has three degrees of freedom: The choice of DDPM or DDIM, the multi-step schedule length and the combination factor $\lambda$.

\section{Experimental results}
\label{sec:ExperimentalResults}

\begin{table}[b]
    \centering
    \resizebox{0.92\linewidth}{!}{
        \begin{tabular}{lrrrrrrr}
            \toprule
                &  &\multicolumn{3}{c}{FFHQ} & \multicolumn{3}{c}{ImageNet} \\
                \cmidrule(lr){3-5} \cmidrule(lr){6-8}
                \textbf{DDPM}&$\lambda$  & PSNR↑  & FID↓ & LPIPS↓ & PSNR↑  & FID↓ & LPIPS↓ \\
            \midrule
                noisy &- & 12.19  & 159.91 & 1.222 & 12.24 & 99.38 & 1.081\\
                one-step&0  & 29.59  & 33.11 & 0.138 & 27.60 & 13.28 & 0.164\\
                6-step  &0 & 28.19 & 22.82 & 0.121  & 26.10 & 9.69 & 0.148\\
                168-step &0 & 26.84  & \textbf{6.77} & \textbf{0.097} & 24.78 & 6.40 & 0.131 \\
                168-step& 0.5 & 29.51 & 16.42 & 0.100 & 25.96 & \textbf{6.34} & \textbf{0.120}\\
                168-step& 0.75 & \textbf{30.03} & 26.86 & 0.123 &  \textbf{27.80} & 8.33 & 0.131\\
            \bottomrule
            \end{tabular}
    } 
    \resizebox{0.92\linewidth}{!}{
        \begin{tabular}{lrrrrrrr}
            \toprule
                &  &\multicolumn{3}{c}{FFHQ} & \multicolumn{3}{c}{ImageNet} \\
                \cmidrule(lr){3-5} \cmidrule(lr){6-8}
                \textbf{DDIM}&$\lambda$  & PSNR↑  & FID↓ & LPIPS↓ & PSNR↑  & FID↓ & LPIPS↓ \\
            \midrule
                noisy &- & 12.19  & 159.91 & 1.222 & 12.24 & 99.38 & 1.081\\
                one-step &0 & 29.59  & 33.11 & 0.138 & 27.60 & 13.28 & 0.164\\
                6-step  &0 & 29.19 & 14.54 & 0.090      & 27.18 & 7.25 & 0.111\\
                168-step &0 & 28.52 & \textbf{5.64} & 0.075 & 26.50 & 5.22 & 0.099 \\
                168-step& 0.2 & 29.59 & 6.69 & \textbf{0.070} &   27.57 & \textbf{5.14} & \textbf{0.091}\\
                168-step& 0.6 & \textbf{30.69} & 18.12 & 0.095 &  \textbf{28.69} & 7.39 & 0.115\\
            \bottomrule
            \end{tabular}
    }
    \caption{Evaluation metrics for the DDIM and DDPM variant of our \Abkuerzung$\;$ method on the FFHQ and ImageNet datasets with noise level $\rho=75$, which leads to a maximal-schedule with $\hat k = 168$. In the second column we depict the scaling factor $\lambda$ of the linear combination. The best method for each performance measure will be depicted in \textbf{bold} letters.}
    \label{tab:DDPMandDDIMDifferentSchedules}
\end{table}

 \begin{figure*}[t]
    \centering
     \begin{minipage}{0.495\textwidth}
         \centering
         \scriptsize
         \subcaption*{DDIM on FFHQ}
         \def\svgwidth{\textwidth}
         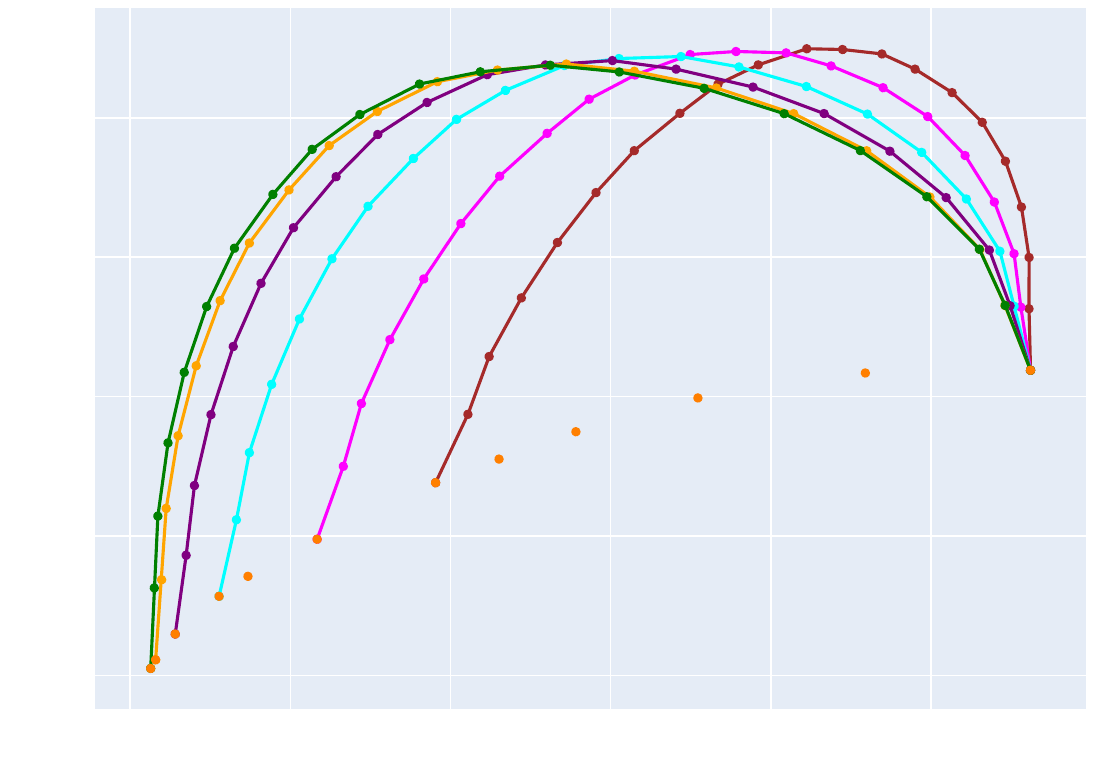
    \end{minipage}
    \begin{minipage}{0.495\textwidth}
        \centering
        \scriptsize
        \subcaption*{DDPM on FFHQ}
        \def\svgwidth{\textwidth}
        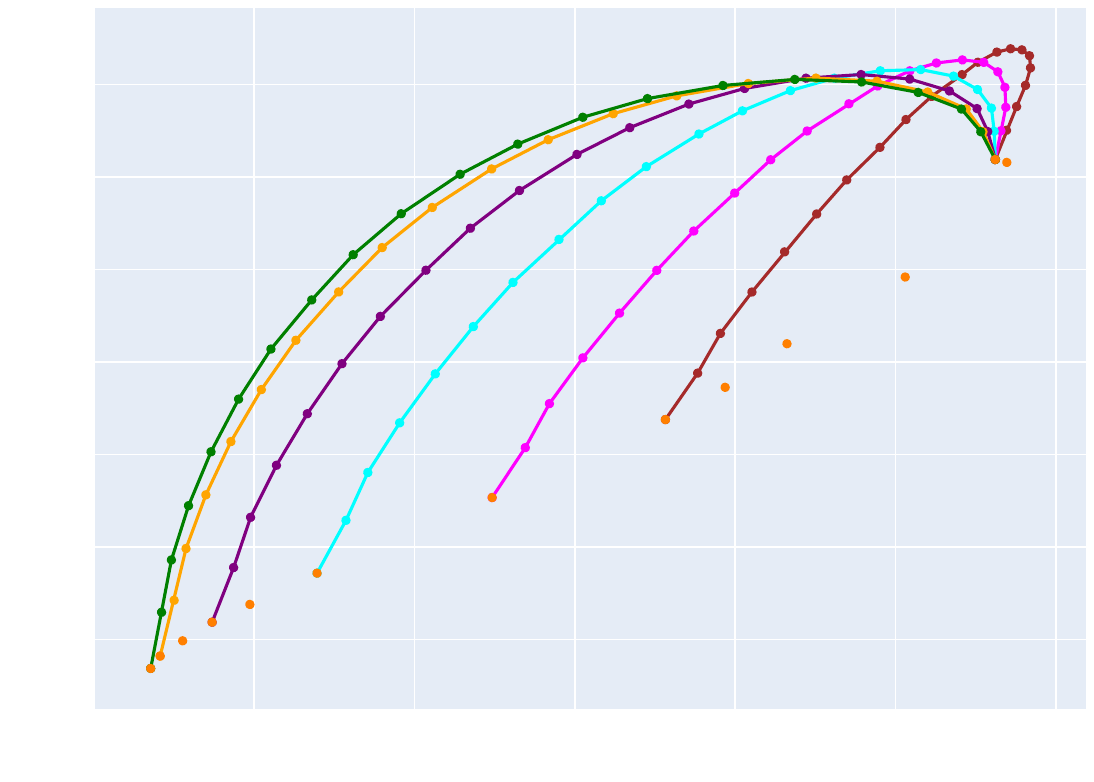
     \end{minipage}
     \begin{minipage}{0.495\textwidth}
         \centering
         \scriptsize
         \subcaption*{DDIM on ImageNet}
         \def\svgwidth{\textwidth}
         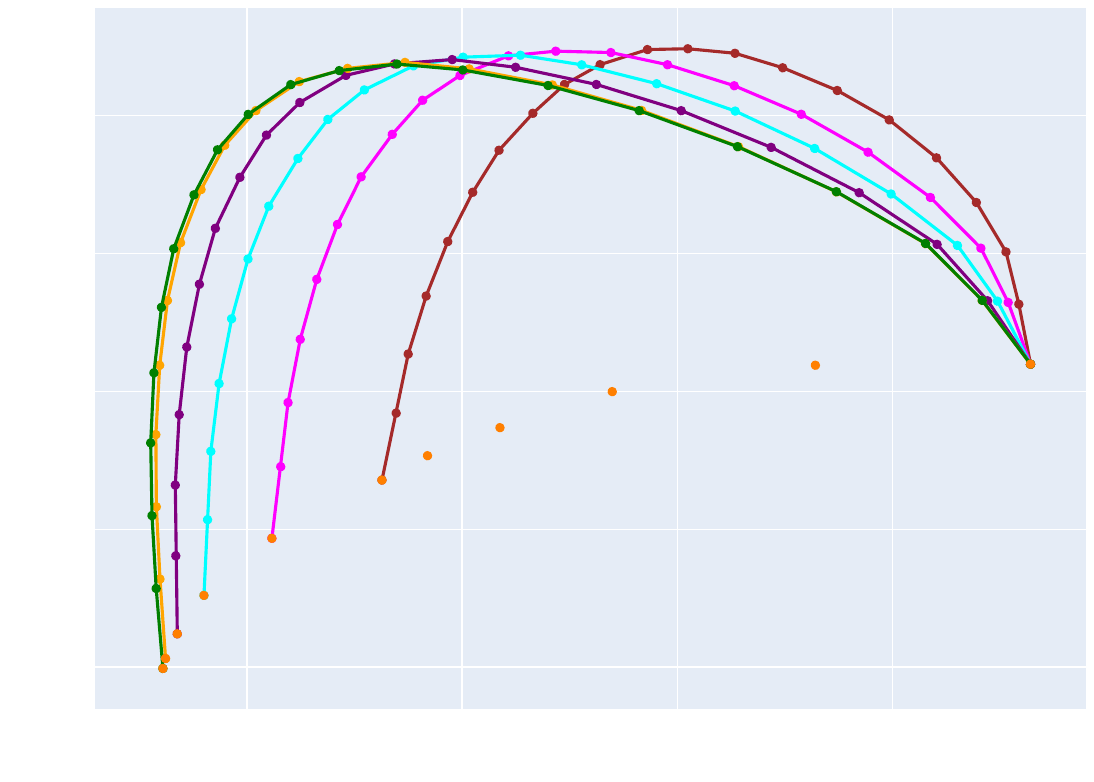
    \end{minipage}%
    \begin{minipage}{0.495\textwidth}
        \centering
        \scriptsize
        \subcaption*{DDPM on ImageNet}
        \def\svgwidth{\textwidth}
        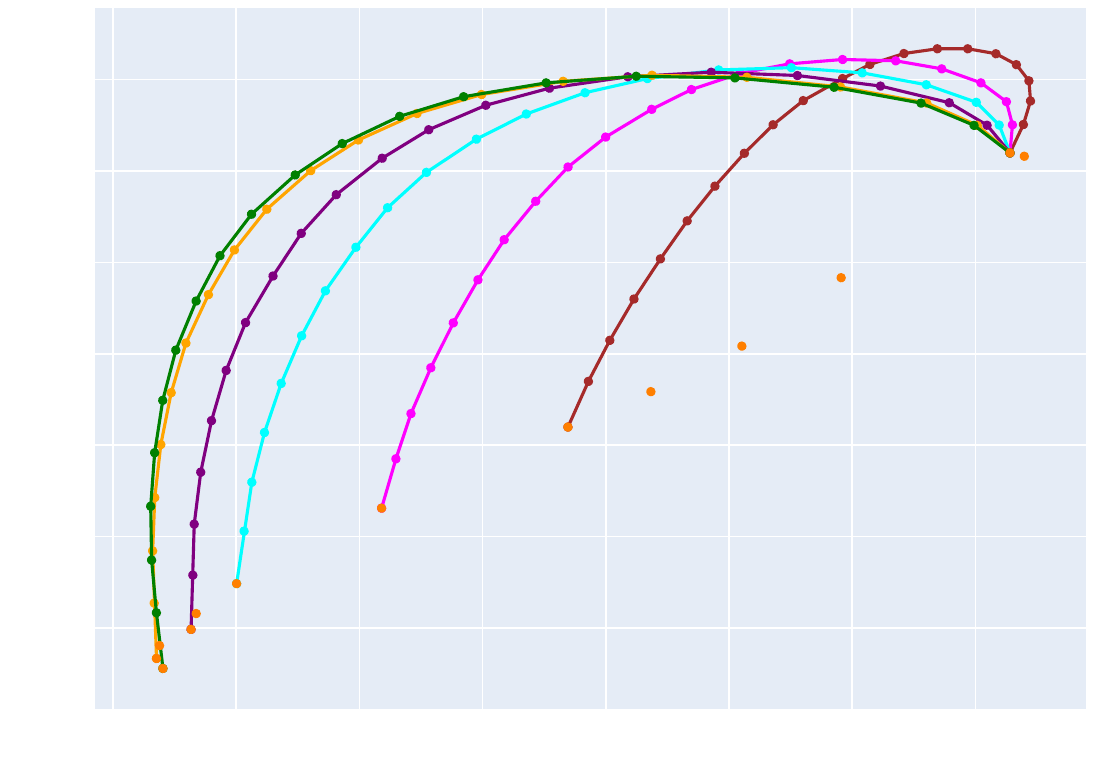
    \end{minipage}
    \caption{Evaluation of our \Abkuerzung$\;$method with different schedule-length and combination factors $\lambda\in[0,1]$ on FFHQ and Imagenet. Both models have been trained on the corresponding dataset. We evaluate the linear combinations for the combination factor $\lambda=0.05k$ with $k=0,1,2,..,20$, \ie all linear combination curves consist of 21 points. In all graphs, the lower, orange string of points correspond to the natural trade-off when varying the schedule length, the length of which is annotated to the points, while the other lines depict the metrics of linear combinations. All test images exhibited a noise level of $ \rho = 75$, \ie we get $\hat k = 168$ with $\hat k$ defined in \cref{subsec:DenosiingWithDiffusionModels}. }
    \label{fig:ImageNetAndFFHQTrade-Off}
\end{figure*}

\begin{figure*}[t]
    \centering
    \begin{minipage}{0.25\textwidth}
         \centering
         \fontsize{6}{5}\selectfont
         \subcaption*{$\rho = 15$}
         \def\svgwidth{0.9\textwidth}
         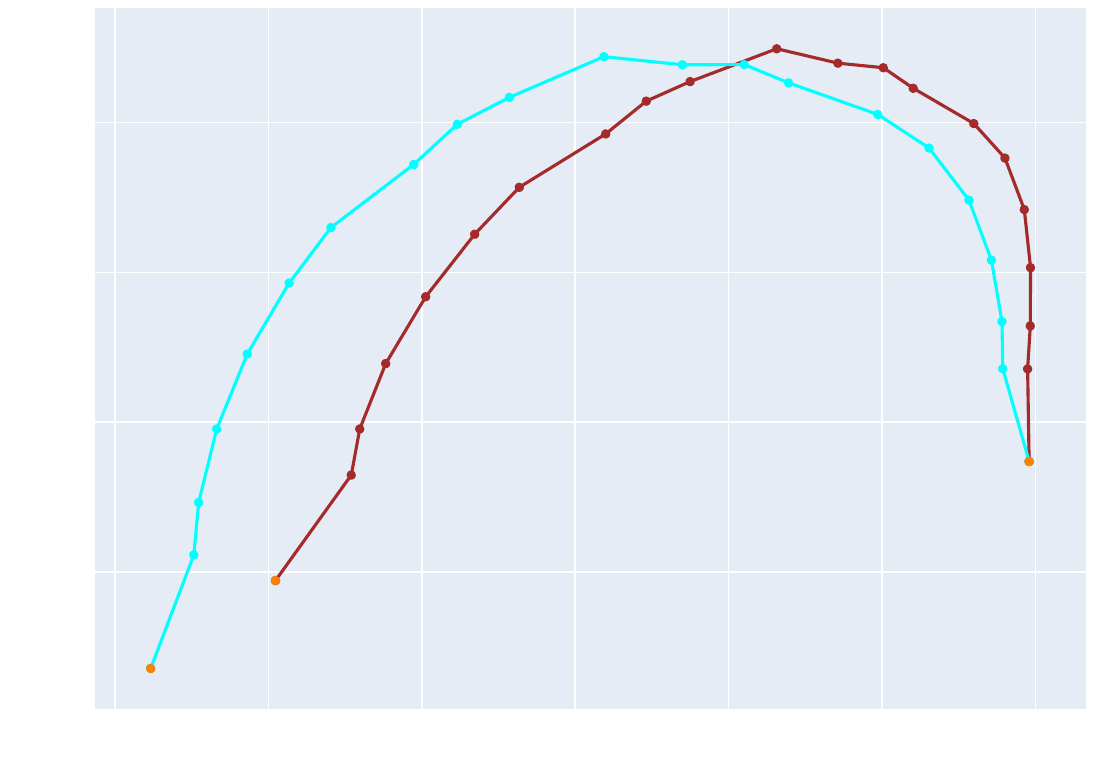
        \end{minipage}%
        \begin{minipage}{0.25\textwidth}
         \centering
         \fontsize{6}{5}\selectfont
         \subcaption*{$\rho = 25$}
         \def\svgwidth{0.9\textwidth}
         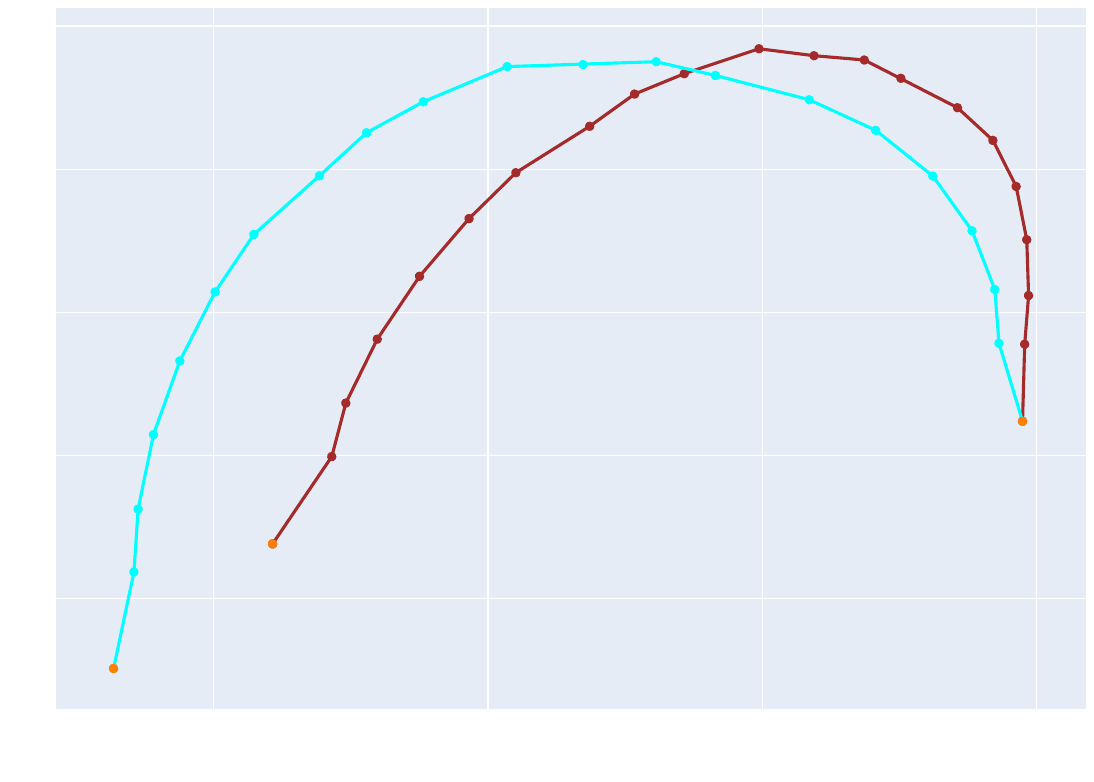
        \end{minipage}%
        \begin{minipage}{0.25\textwidth}
         \centering
         \fontsize{6}{5}\selectfont
         \subcaption*{$\rho = 50$}
         \def\svgwidth{0.9\textwidth}
         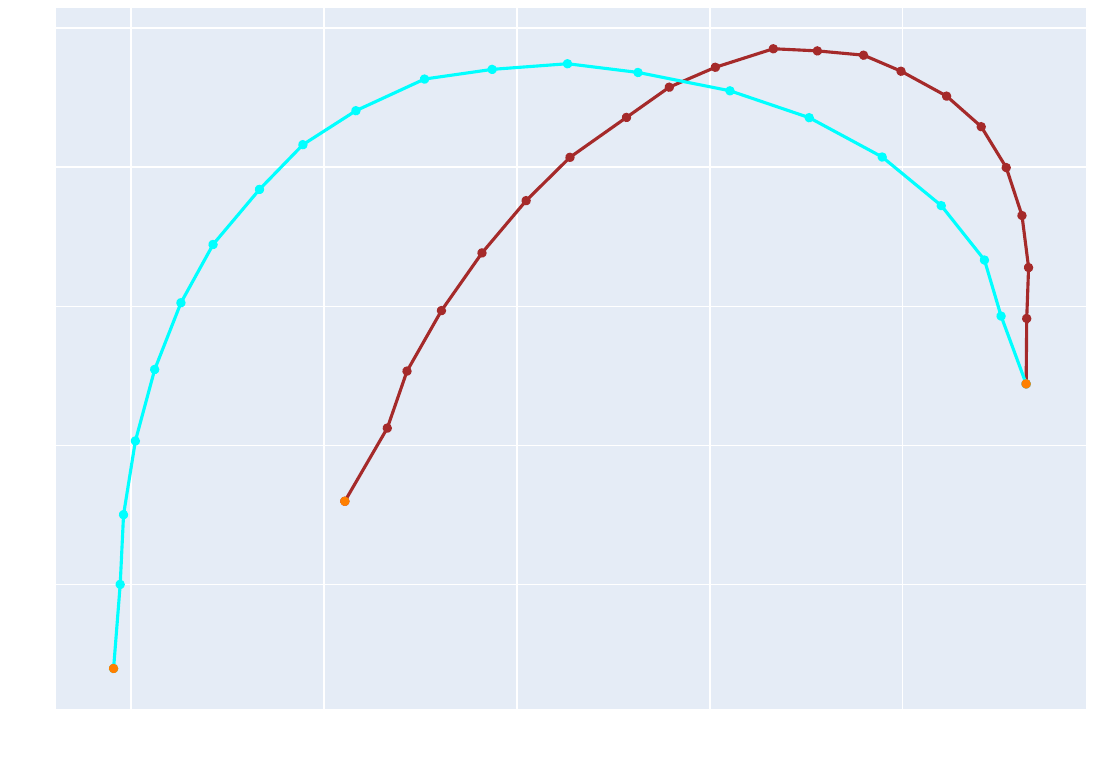
        \end{minipage}%

    \caption{Evaluation of the DDIM variant on the FFHQ dataset for different noise scales $\rho$.}
    \label{fig:DifferentNoiseLevels}
\end{figure*}

 \begin{figure*}[t]
  \centering
    \begin{minipage}{0.5\textwidth}
        \centering
        \scriptsize
        \subcaption*{BSD68}
        \def\svgwidth{\linewidth}
        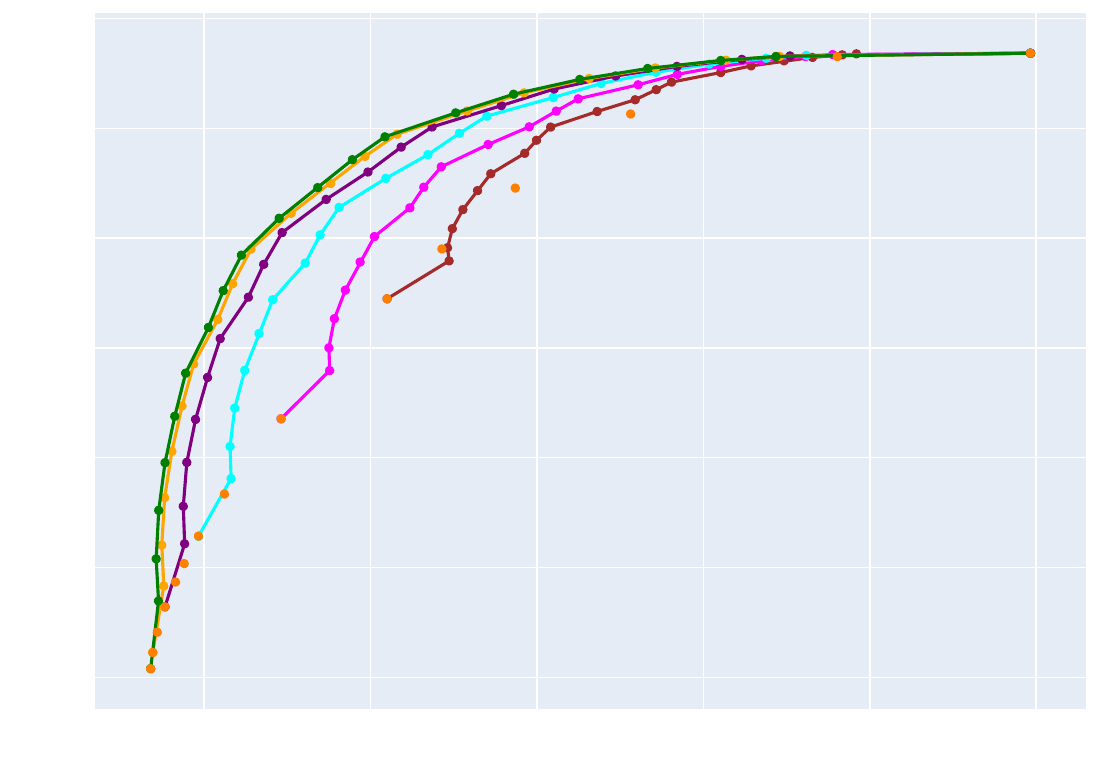
    \end{minipage}%
    \begin{minipage}{0.5\textwidth}
        \centering
        \scriptsize
        \subcaption*{ImageNet}
        \def\svgwidth{\linewidth}
        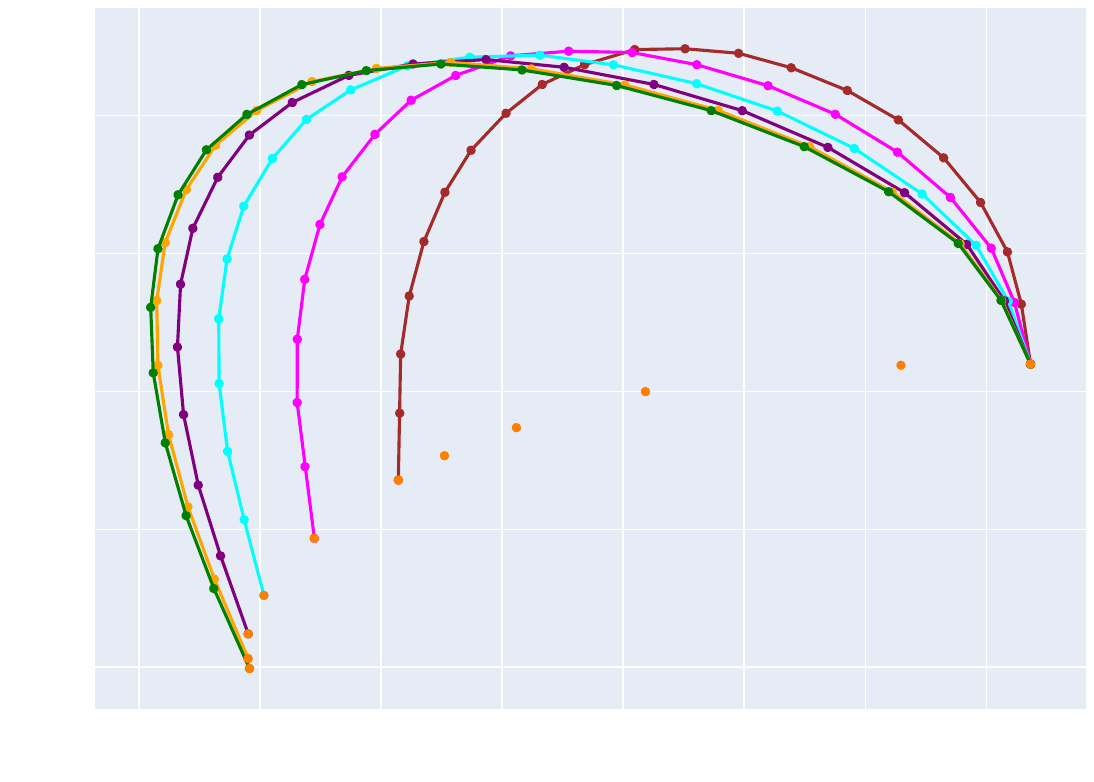
     \end{minipage}

   \caption{Evaluation of our \Abkuerzung$\;$method in the DDIM variant with varying schedule lengths and combination factors $\lambda \in [0,1]$ on BSD68 and ImageNet. The setup follows that of \cref{fig:ImageNetAndFFHQTrade-Off}, but with some differences in the reported metrics. The noise level was set to $\rho = 75$.}
   \label{fig:BSD68Trade-Off}
\end{figure*}

\begin{table*}[tp]
    \centering
    \resizebox{\linewidth}{!}{%
    \begin{tabular}{|c|c||c|c|c|c||c|c|c|c||c|c|c|c|}
        \hline  
        &&\multicolumn{4}{c||}{BSD68} &\multicolumn{4}{c||}{Kodak24 } &\multicolumn{4}{c|}{McMaster}\\
        && PSNR↑ & LPIPS↓   & FID↓ &NIQE↓ & PSNR↑  & LPIPS↓ & FID↓ &NIQE↓ & PSNR↑ & LPIPS↓ & FID↓ &NIQE↓  \\ \hline
        \parbox[t]{2mm}{\multirow{7}{*}{\rotatebox[origin=c]{90}{$\rho=15$}}}&Restormer \cite{Zamir2021_Restormer}& 34.40	&0.054	&18.30	&3.85	&35.35	&0.066	&23.64	&3.76	&35.61	&0.049	&35.65	&4.26\\ 
        &MambaIR \cite{Guo2024_MambaIR}           & \psnr{34.48}&0.052	        &16.39       &    3.78     &   35.42	&0.064	       &   21.91   &  3.66       &    35.70   &     0.048	&34.00	    &   4.18  \\ 
        &DOT Restormer \cite{adrai2024DeepOptimal}& 26.74       &0.147          &21.77       &    4.29     &   28.46    &0.131         &   29.69   &  4.09       &    28.92   &     0.114   &43.26      &   4.33  \\  
        &\Abkuerzung $\;\lambda=0$                & 33.92 	    &\lpips{0.028}	&\fid{6.45}  &\niqe{3.25}  &	34.78   & 0.030	       &\fid{10.05}&  \niqe{3.14}& 	  35.06   & 	0.020	&\fid{11.37}&\niqe{3.51}\\
        &\Abkuerzung $\;\lambda=0.2$              & 34.11 	    &0.030          &7.15        &	3.37       &	39.27   & \lpips{0.008}&   10.84   &  3.17       &    39.00   &\lpips{0.004}&14.01      &  	3.56   \\ 
        &\Abkuerzung $\;\lambda=0.6$              & 34.37 	    &0.037	        &9.83        &	3.53       &\psnr{39.50}& 0.009        &   14.39   &  3.43       &\psnr{39.21}&  	0.005	&20.76      &  	3.90\\
        & \Abkuerzung $\;\lambda=1$               & \psnr{34.48}&0.051	        &16.26       &	3.83       &	35.42   & 0.065	       &   22.77   &  3.73       &    35.73   &     0.048	&33.55      &   4.24  \\
        \hline
        \hline
        \parbox[t]{2mm}{\multirow{7}{*}{\rotatebox[origin=c]{90}{$\rho=25$}}}&Restormer \cite{Zamir2021_Restormer}& 31.79	&0.094&	26.68	&4.15	&32.93	&0.106	&32.13	&4.09	&33.34	&0.078	&47.42	&4.70\\ 
        &MambaIR\cite{Guo2024_MambaIR}            & 31.86	    &0.093	      &24.83	 &4.06	     &32.99	      &0.104	    &30.93	    &3.95	    &33.43	     &0.077	       &46.12	   &4.62  \\ 
        &DOT Restormer \cite{adrai2024DeepOptimal}& 26.60       &0.137        &21.46     &4.13       &28.04       &0.136        &28.93      &4.01       &28.96       &0.102        &40.96      &4.17\\  
        &\Abkuerzung $\;\lambda=0$                & 31.13	    &\lpips{0.047}&\fid{7.89}&\niqe{3.20}&32.15	      &0.052	    &12.93	    &\niqe{3.07}&32.59	     &0.034	       &\fid{13.89}&\niqe{3.49}\\ 
        &\Abkuerzung $\;\lambda=0.2$              & 31.38	    &0.048	      &8.24	     &3.32	     &36.35	      &\lpips{0.014}&\fid{12.87}&3.11	    &36.14	     &\lpips{0.008}&15.74	   &3.54  \\ 
        &\Abkuerzung $\;\lambda=0.6$              & 31.75	    &0.060	      &11.82	 &3.58	     &\psnr{36.67}&0.018	    &16.30	    &3.45	    &\psnr{36.44}&0.010	       &24.04	   &3.96\\
        &\Abkuerzung $\;\lambda=1$                & \psnr{31.90}&0.086	      &21.66	 &4.05	     &33.03	      &0.097	    &27.46	    &3.92	    &33.50	     &0.071	       &41.10	   &4.49 \\
        \hline
        \hline
        \parbox[t]{2mm}{\multirow{7}{*}{\rotatebox[origin=c]{90}{$\rho=50$}}}&Restormer \cite{Zamir2021_Restormer}& 28.60	&0.179	&41.58	&4.63	&29.87	&0.183	&45.41	&4.66	&30.30	&0.135	&62.13	&5.39\\ 
        &MambaIR\cite{Guo2024_MambaIR}           & 28.67	  &0.179	    &38.85	    &4.52	    &29.92	     &0.180	       &43.38	   &4.40	   &30.35	    &0.134	      &61.28	    &5.20  \\ 
        &DOT Restormer\cite{adrai2024DeepOptimal}& 25.55      &0.163        &24.89      &4.20       &26.94       &0.167        &32.53      &4.01       &27.67       &0.120        &41.17        &4.06\\  
        &\Abkuerzung $\;\lambda=0$               & 27.75	  &0.093	    &\fid{10.95}&\niqe{3.18}&28.88	     &0.104	       &19.51	   &\niqe{2.92}&29.37	    &0.071	      &\fid{20.10}	&\niqe{3.30}\\ 
        &\Abkuerzung $\;\lambda=0.2$             & 28.09	  &\lpips{0.092}&11.17	    &3.29	    &32.59	     &\lpips{0.035}&\fid{18.38}&3.01	   &32.34	    &\lpips{0.021}&20.58	    &3.42  \\ 
        &\Abkuerzung $\;\lambda=0.6$             & 28.58	  &0.110	    &15.81	    &3.62	    &\psnr{33.02}&0.041	       &20.79	   &3.42	   &\psnr{32.73}&0.025        &28.39	    &3.99  \\
        &\Abkuerzung $\;\lambda=1$               &\psnr{28.77}&0.161	    &31.68	    &4.32	    &30.02	     &0.162	       &35.64	   &4.20	   &30.50	    &0.121	      &52.19	    &4.84  \\
        \hline
        
    \end{tabular}%
    }
    \caption{Comparison between state of the art image denoising methods and our method using the maximal-schedule. We highlight the best PSNR performances in \psnr{red}, best LPIPS performances in \lpips{blue}, best FID performances in \fid{green}, and best NIQE performances in \niqe{yellow}.}
    \label{table:SOTAComparison}
\end{table*}

\begin{figure*}[t]

     \centering
     \includegraphics[width=\textwidth]{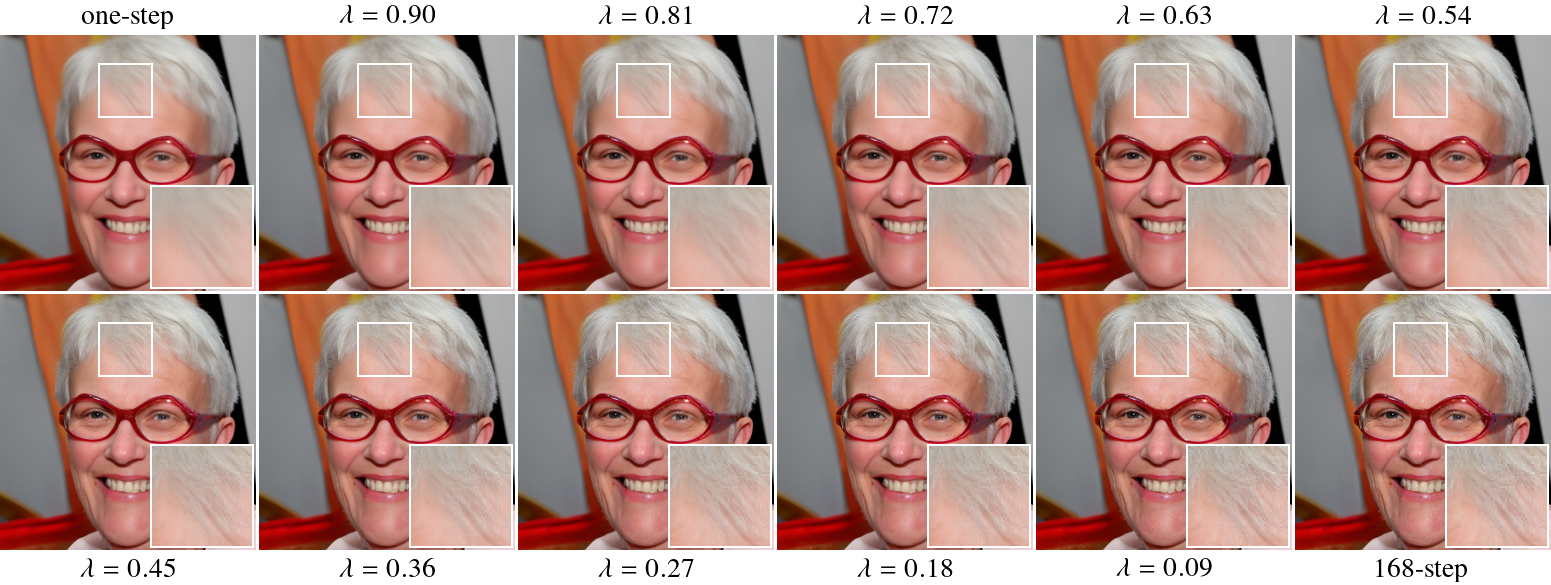}
     \caption{An example of the linear combination of the one-step schedule and the 168-step schedule of the DDIM variant while denoising an image taken from the FFHQ dataset with noise level $\rho = 75$. In the upper left we show the one-step schedule, while in the bottom right we show the 168-step schedule. All intermediate pictures from left to right and top to bottom are linear combinations. The noisy image and ground truth can be found in  \cref{fig:ExampleDenoisedImages}.}
     \label{fig:ExampleLinearCombinations}
\end{figure*}

\begin{figure*}[t]

     \centering
     \includegraphics[width=\textwidth]{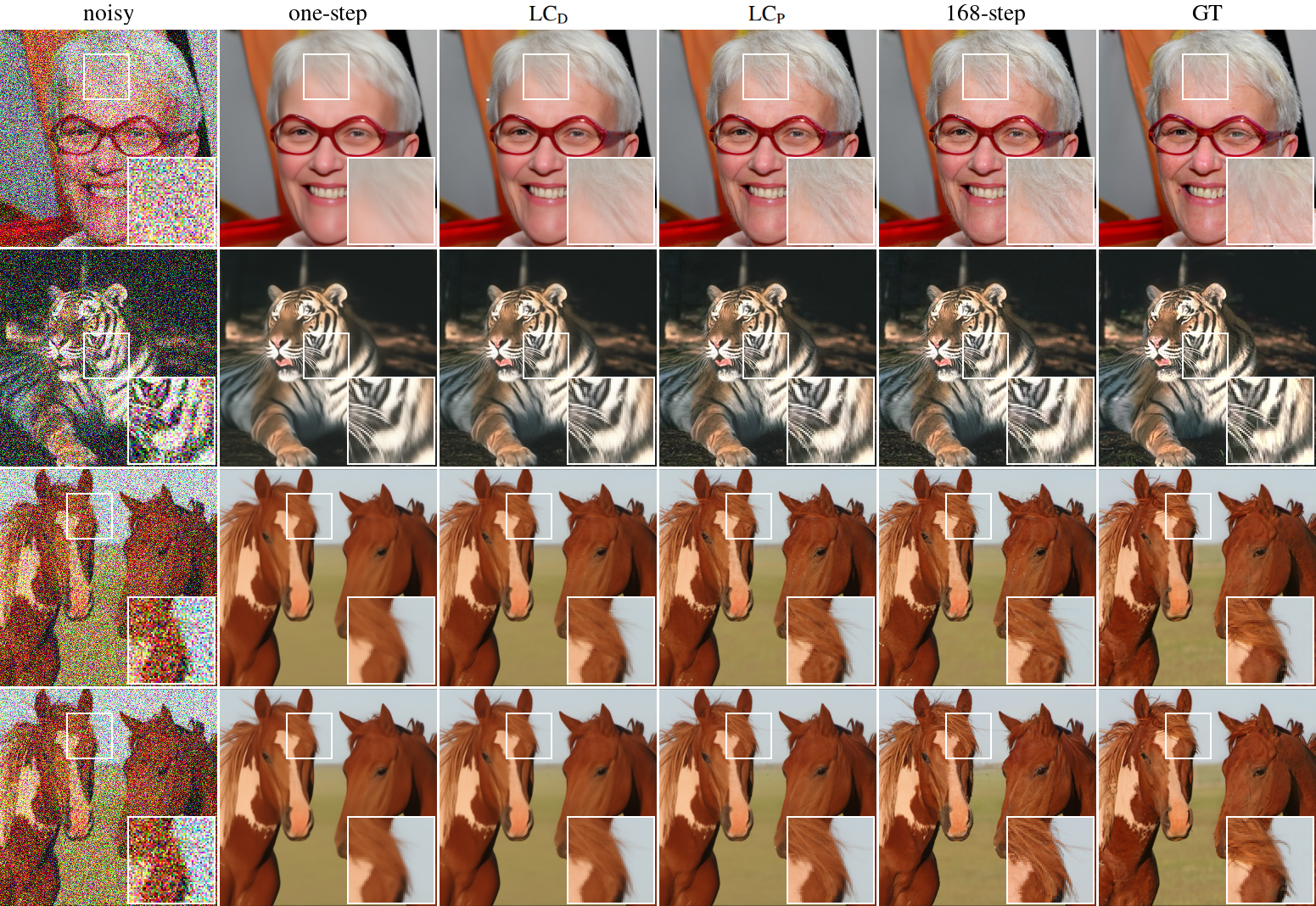}
     \caption{Denoised images using diffusion models with different inference lengths and our proposed linear combination method. The first three rows presents the DDIM variant on the FFHQ, BSD68 and ImageNet dataset, respectively. The fourth row displays the DDPM variant with an ImageNet sample. The first column contains the noisy input images, all with a noise level of $\rho = 75$, corresponding to $\hat{k} = 168$. The column labeled $\text{LC}_\text{D}$ represents a linear combination of the 168-step schedule with emphasis on distortion, where we set the combination factor to $\lambda=0.6$ for DDIM and $\lambda=0.75$ for DDPM. The column $\text{LC}_\text{P}$ highlights a linear combination of the 168-step focused on perception, using $\lambda=0.2$ for DDIM and $\lambda=0.5$ for DDPM. The exact performance metrics for these denoising schemes can be found in \cref{tab:DDPMandDDIMDifferentSchedules}.}
     \label{fig:ExampleDenoisedImages}
\end{figure*}

As datasets we employ the well-known ImageNet dataset \cite{deng2009_imagenet} and the Flickr-Faces-HQ (FFHQ) dataset \cite{karras2021_stylegan_ffhq}. In order to compare to other methods, we evaluate the performance of our algorithm on the datasets Kodak24 \cite{franzen1999_kodak24}, BSD68 \cite{martin2001_bsd68}, and McMaster \cite{zhang2011_mcmaster}. 

In all experiments, predefined random subsets of these datasets are used. For instance, the same 1000-image subset of the ImageNet training set is utilized across all evaluations. Images are cropped and resized to 256$\times$256 following the procedure described in \cite{Dhariwal2021_diff_beat_gan}. We use the inference process described in  \cref{subsec:DenosiingWithDiffusionModels} together with pretrained diffusion models. All networks used were trained using the framework provided by \cite{Dhariwal2021_diff_beat_gan}, which implements the DDPM and DDIM models. Notably, DDPM and DDIM share the same foundational model, allowing a single trained model to support both sampling methods. Further performance gains are achieved through modifications to the neural network architecture. We employ the model trained on the ImageNet dataset at 256$\times$256 resolution, as well as a model trained on the FFHQ dataset resized to 256$\times$256, as reported in \cite{chung2022_DPS_ffhq_checkpoint}. Kodak24 and McMaster images are uniformly sized at 500$\times$500 pixels, while BSD68 images have resolutions of either 481$\times$321 or 321$\times$481 pixels. Following the widely used KAIR benchmark\footnote{https://github.com/cszn/KAIR}, we evaluate the model on four different 256$\times$256 tiles of each image and report the average results.

To evaluate perceptual quality we use the Fréchet inception distance (FID)in the widely adopted \texttt{pytorch-fid} \cite{Seitzer2020FID} implementation. 
To lower the computational cost, we compute the FID score on a subset of 5000 test samples. 
Clean images are first degraded with noise and then denoised using the proposed methods. 
The resulting denoised images are compared to the clean images using the FID metric. 
For FID evaluation on the BSD68 dataset, where the number of available images is significantly smaller, we extract 5000 patches of size 64$\times$64 from each test image and compute the FID
score by comparing the clean and denoised patches. 
We use Learned Perceptual Image Patch Similarity (LPIPS) \cite{zhang2018unreasonable} and Natural Image Quality Evaluator (NIQE) \cite{mittal2013making} as a further visual quality metrics, and use the original code provided by its authors, with AlexNet as the encoder for the latter.
To measure the distortion we use PSNR as implemented in the KAIR codebase.

In \cref{table:SOTAComparison}, we compare our approach against three state-of-the-art methods: two traditional denoisers, Restormer \cite{Zamir2021_Restormer} and MambaIR \cite{Guo2024_MambaIR}, both of which perform exceptionally well in the distortion-focused regime, and DOT Restormer \cite{adrai2024DeepOptimal}, which serves as a benchmark for perceptual quality. For DOT Restormer, we use the scaling factor $\hat{x}_{1.7}$, as it achieves the best reported perceptual results for Gaussian denoising, according to \cite[Table 1]{adrai2024DeepOptimal}.

\subsection{Quantitative results}
\cref{tab:DDPMandDDIMDifferentSchedules}, \cref{fig:ImageNetAndFFHQTrade-Off}, and \cref{fig:BSD68Trade-Off} illustrate the impact of different inference schedules on both the DDPM and DDIM variants, comparing their performance without any linear combination (\ie $\lambda = 0$) to that achieved with linear combinations. For both variants, shorter schedules yield better distortion results, while longer schedules trade distortion for improved perceptual quality, and the linear combinations significantly enhance the distortion-perception trade-off. Furthermore, \cref{fig:ImageNetAndFFHQTrade-Off}, \cref{fig:DifferentNoiseLevels}, and \cref{fig:BSD68Trade-Off} demonstrate that our method consistently achieves a favorable distortion-perception trade-off, regardless of the scaling factor $\lambda \in [0,1]$, the noise level $\rho$, the employed metrics, test-dataset, or the specific perception-focused inference schedule used in the linear combination. This observation aligns with theoretical predictions by Fridirich \etal \cite{Freirich2021_distortion-perception_tradeoff_wasserstein}. As reported in \cref{table:SOTAComparison}, our method surpasses current state-of-the-art approaches in both distortion and perception when using the DDIM variant and achieves a favorable trade-off. For instance, in the BSD68 case with $\rho=15$, our method \Abkuerzung$\; \lambda=0.6$ sacrifices only $0.11$ PSNR compared to MambaIR while reducing its FID by half, and in the McMaster case with $\rho=50$, our method \Abkuerzung$\; \lambda=0.2$ increases PSNR by $2$ compared to MambaIR, while sanctimoniously reducing its FID by two thirds. Note that our methods use only one neural network, the DDPM one, while other methods use different models for different noise levels. However, we use a model that is trained on significantly more data, and has much larger capacity compared to the other methods. Moreover, \cref{table:SOTAComparison}, \cref{fig:ImageNetAndFFHQTrade-Off} and \cref{fig:BSD68Trade-Off} indicate that, for certain scenarios, specific scaling factors $\lambda \in [0,1]$ allow the linear combination to enhance the performance of initial denoisers in PSNR, LPIPS, or FID.

\subsection{Qualatative results}
\cref{fig:ExampleDenoisedImages} and \cref{fig:ExampleLinearCombinations} show that our approach of denoising images can produce high quality outputs. The advantages of the different outputs are particularly discernible in highly detailed areas, such as the hair. In \cref{fig:ExampleDenoisedImages}, the maximal-step schedule shows images that appear quite natural, but some diversions to the ground truth are visible. The one-step schedule appears quite smoothed, which seems unnatural, though it shows fewer deviations to the ground truth as the maximal-step schedule. In some cases, $\text{LC}_\text{D}$ shows even better resemblance to the ground truth, while $\text{LC}_\text{P}$ shows both good similarity to the ground truth and natural looking features. \cref{fig:ExampleLinearCombinations} illustrates the transition from the one-step to the maximal-step schedule via a linear combination, highlighting the advantages of different combination factors.

\section{Conclusion}
We introduced the \Name$\;$(\Abkuerzung), a novel denoising method using the inherent iterative denoising structure of diffusion models. It achieves advantageous distortion-perception trade-offs, and its focus can be shifted through a simple hyperparameter adjustment. We demonstrated that our method achieves state-of-the-art performance, and that the beneficial trade-offs are achieved consistently.

\newpage
{
    \small
    \bibliographystyle{ieeenat_fullname}
    \bibliography{main}
}

\end{document}